\documentclass[letterpaper, 10 pt, conference]{ieeeconf}  

\IEEEoverridecommandlockouts                              

\usepackage{graphicx} 
\graphicspath{{./pdf/}} 
\DeclareGraphicsExtensions{.pdf,jpg}
\usepackage[export]{adjustbox} 
\usepackage[skip=1pt,labelformat=empty]{subcaption} 
\usepackage[cmex10]{amsmath} 
\usepackage[short]{optidef} 
\usepackage[ruled,linesnumbered]{algorithm2e} 
\usepackage{caption} 
\usepackage{bm} 
\usepackage{epsfig} 
\usepackage{times} 
\usepackage{color} 
\usepackage{amssymb}  
\usepackage{setspace}

\newcommand{\mytilde}{\raise.17ex\hbox{$\scriptstyle\mathtt{‌​\sim}$}}
\usepackage{xparse,soul}
\usepackage{multirow}
\usepackage{cite}
\usepackage[font=small,labelfont=bf]{caption}

\usepackage[utf8]{inputenc}
\usepackage{mathtools}
\DeclareMathOperator*{\argmax}{arg\,max}

\title{\LARGE \bf
Simultaneous Estimation of Shape and Force along Highly Deformable Surgical Manipulators Using Sparse FBG Measurement
}

\author{Yiang Lu$^{1}$\textsuperscript{\textdagger}, Bin Li$^{1}$\textsuperscript{\textdagger}, Wei Chen$^{1}$, Junyan Yan$^{1}$, Shing Shin Cheng$^{1}$, Jiangliu Wang$^{1}$,  
Jianshu Zhou$^{1,3}$, \\
Qi Dou$^{2,3}$, and Yun-Hui Liu$^{1,3}$
\thanks{This work is supported in part by Shenzhen Portion of Shenzhen-Hong Kong Science and Technology Innovation Cooperation Zone under HZQB-KCZYB-20200089, in part by the Research Grants Council of Hong Kong under Grant T42-409/18-R, Grant 14202918, Grant 14207119, Grant 14207320, and Grant 14207423, in part by the Hong Kong Centre for Logistics Robotics, in part by the Multi-Scale Medical Robotics Centre, InnoHK, and in part by the VC Fund 4930745 of the CUHK T Stone Robotics Institute. \textit{(Corresponding author: Yun-Hui Liu and Jianshu Zhou.)}}%
\thanks{$^{1}$T Stone Robotics Institute, Department of Mechanical and Automation Engineering, The Chinese University of Hong Kong, Hong Kong.}
\thanks{$^{2}$Department of Computer Science and Engineering, The Chinese University of Hong Kong, Hong Kong.}
\thanks{$^{3}$Hong Kong Center for Logistics Robotics, Hong Kong.}
\thanks{\textsuperscript{\textdagger}Y. Lu and B. Li contributed equally to this work.}
}

\begin{document}

\maketitle
\thispagestyle{empty}
\pagestyle{empty}

\begin{abstract}
Recently, fiber optic sensors such as fiber Bragg gratings (FBGs) have been widely investigated for shape reconstruction and force estimation of flexible surgical robots. 
However, most existing approaches need precise model parameters of FBGs inside the fiber and their alignments with the flexible robots for accurate sensing results.
Another challenge lies in online acquiring external forces at arbitrary locations along the flexible robots, which is highly required when with large deflections in robotic surgery.
In this paper, we propose a novel data-driven paradigm for simultaneous estimation of shape and force along highly deformable flexible robots by using sparse strain measurement from a single-core FBG fiber.
A thin-walled soft sensing tube helically embedded with FBG sensors is designed for a robotic-assisted flexible ureteroscope with large deflection up to 270$^\circ$ and a bend radius under 10 mm.
We introduce and study three learning models by incorporating spatial strain encoders, and compare their performances in both free space without interactions as well as constrained environments with contact forces at different locations.
The experimental results in terms of dynamic shape-force sensing accuracy demonstrate the effectiveness and superiority of the proposed methods.
\end{abstract}

\section{INTRODUCTION}

With development of flexible surgical robots endowing enhanced reach and dexterity, they have been increasingly utilized in minimally invasive surgery (MIS) as well as natural orifice transluminal endoscopic surgery (NOTES) \cite{kim2022advancement,yan2022continuum}.
To guarantee precise control and stable manipulation of these flexible endoscopes and instruments, real-time perception of their shape is crucial as accurate feedback and guidance for both teleoperation and autonomous tasks \cite{10161505}.
Moreover, contact force sensing at arbitrary locations along the flexible robot is still challenging for safety-critical surgical interventions.
It is critical to acquire both magnitude and location of external force applied on the robot through tortuous paths in unstructured environments, especially for those with large deflections and curvatures, such as flexible ureteroscopes and colonoscopes \cite{alian2022current}.


Numerous sensing techniques were investigated for force or shape sensing of flexible manipulators, including model-based approaches, electromagnetic tracking, and intraoperative imaging modalities (e.g., fluoroscopy and ultrasound).
However, they suffer from limitations associated with their working principles, making it challenging to deploy these systems for specific surgical procedures \cite{shi2016shape,lu2021robust,zhang2022shape}.
In recent years, fiber Bragg grating (FBG) sensors have become popular in flexible robotic surgery \cite{li2019distributed,zhang2023fiber}, due to their benefits of compact size, high frequency, and biocompatibility.
By grouping multiple single-core FBG fibers in the straight configuration aligning paralleled with the centroidal axis \cite{deaton2023towards}, or by twisting the fibers in the helical configuration \cite{wei2017novel}, the flexible robot shape can be reconstructed.
Besides, multi-core optical fiber with FBGs based on optical frequency-domain reflectometry (OFDR) is an alternative way to simplify the integration and alleviate the alignment difficulties \cite{donder2021kalman,lu2023robust,wang2023fast,ha2023sensor,zhang2023vascular,lu2024adaptive}.
The fiber with multiple twisted cores has also been studied to measure twists accurately \cite{khan2021curvature, yang2022three}.




Apart from shape sensing, FBG sensors in both straight and helical configurations have been developed for contact force estimation during flexible robotic surgery \cite{xu2016curvature,li2018three,jiang2023fiber}.
Kinematics and mechanics models such as Cosserat rod theory \cite{khan2017force, qiao2021force,deaton2023simultaneous} were employed to estimate forces based on the prior knowledge of robot configuration.
However, most existing works focus on sensing the external force applying to the distal tip of continuum robots or assuming that the contact location is known.
Localization of arbitrary contact force along the flexible robot together with its magnitude estimation has shown the possibility from the robot shape in combination with Cosserat rod model \cite{qiao2021force,al2021fbg}.
This is highly demanded for highly deformable continuum robots in unknown scenes, because the surrounding anatomy can interact with the robot everywhere.
Besides, the performances of the model-based approaches suffer from inaccurate characteristic parameters like alignment of FBGs with the robot, especially for those adopting the fiber helically wound, which require engraving precision grooves for FBG integration.
Online calibrations of both FBG sensing units and deformation model of continuum robots have been proposed, but they require fusion with an external imaging modality \cite{alambeigi2019scade}.


Machine learning methods have gained great attention for discovering the representation of highly nonlinear behavior, which can deal with the limitations discussed above, and can also account for increasing sensory noises due to spectral distortion of FBG fiber when bending large deformation.
Amongst, neural networks (NNs) were commonly utilized for shape sensing of flexible robots \cite{wang2020eye,sefati2020data}.
Ha \textit{et al.} \cite{ha2022shape} learned the curvature and twist for shape reconstruction of the continuum robot based on multi-core FBGs placed off-center to the robot, thus alleviating precise knowledge of fiber location.
Lately, they developed a network to localize the contact point of external force along the robot \cite{ha2022contact}.
Long short-term memory (LSTM) network, as a classical temporal NN, was deployed to handle the hysteresis issue of a catheter system \cite{wu2021hysteresis}.
Hao \textit{et al.} \cite{hao2023two} implemented an LSTM network for shape sensing and distal force estimation of a continuum manipulator.
In addition to placing FBGs in straight configurations, Wang \textit{et al.} \cite{wang2023learning} demonstrated the possibility of estimating the pose of continuum robots, by fusing eye-in-hand endoscopic vision with sparse strain measurement from helically wrapped FBG sensors in one single-core fiber. 
However, most studies were only validated with fixed curvature lacking actuation, or small deformations of the robots, while neglecting those with large bending angles required for ureteroscopy, colonoscopy, and choledochoscopy.
Nevertheless, shape estimation of a flexible robot with bending up to 320$^\circ$ was investigated based on an NN model combining FBG signals with actuation data \cite{hao20232d}.


Targeting these limitations, we propose a new data-driven paradigm for simultaneous shape-force estimation of flexible surgical robots by using sparse FBG measurement and taking advantage of continuous shape transformations in spatial domain.
A soft sensing tube is designed with helically wrapped FBGs on a single-core optical fiber, which is bio-compatible, low-cost, and capable of large deformation.
Both shape and contact location/magnitude of force applied along the robot can be determined by comparing three different supervised learning algorithms with spatial strain encoders, which can alleviate the requirement for precise FBG placement on the tube.
We fabricate and integrate the tube onto a robotic-assisted flexible ureteroscope (RAFUS), where comprehensive experiments have been performed to evaluate the proposed learning models.
We summarize the main contributions of this paper as follows:
\begin{enumerate}
    \item Learning-based estimation methods by leveraging data exploration of sparse FBG strains are proposed to not only acquire the shape but also simultaneously estimate the force for the flexible surgical robot.
    \item The proposed networks can learn both the magnitude and contact position of external force at arbitrary locations along the flexible robot.
    \item A soft sensing tube with a helically wound single-core FBG fiber is newly designed and implemented for the RAFUS allowing large deflection ratio.
    \item A robotic platform is integrated for efficient data collection. Experiments are performed to compare the proposed approaches in dynamic conditions, and the results demonstrate their feasibility and superiority.
\end{enumerate}

\begin{figure}[!ht]
  \centering
  \includegraphics[width=\linewidth]{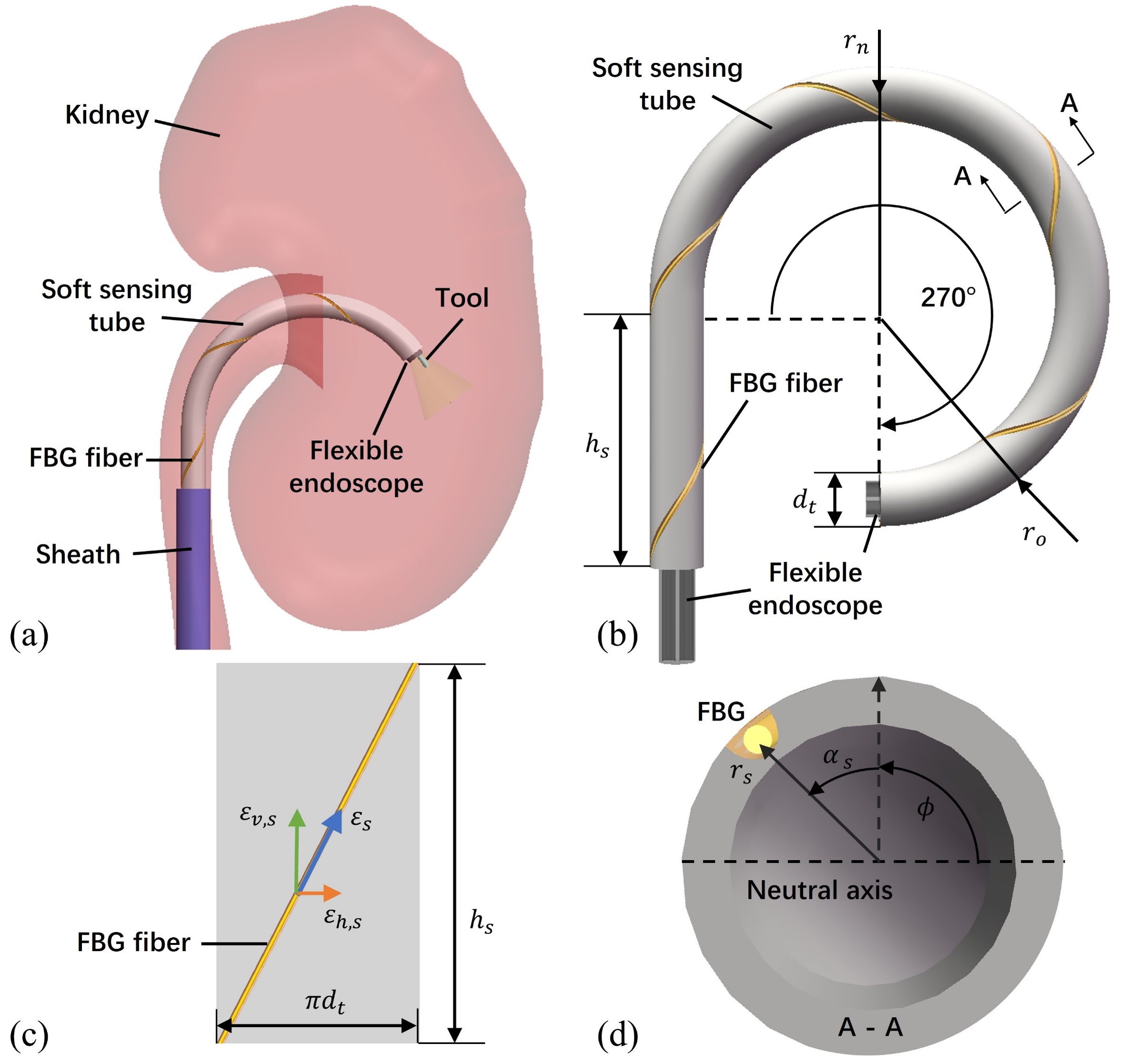}
  \caption{Design of FBG-based soft sensing tube. (a) Conceptual representation of flexible endoscope with FBG sensing in kidney; (b) Soft tube with a helically wrapped single-core FBG fiber; (c) FBG strain decomposition; (d) Cross-section diagram of the tube.}
  \label{fig:diagram}
  \vspace{-0.4cm}
\end{figure}

\section{Design of FBG-Based Soft Sensing Tube}
In this section, we introduce the design of an FBG-based soft tube, and then derive a model-based sensing method to compare with our learning-based approaches, for shape and force estimation of flexible surgical robots.

\begin{figure*}[!ht]
  \centering
  \includegraphics[width=0.9\linewidth]{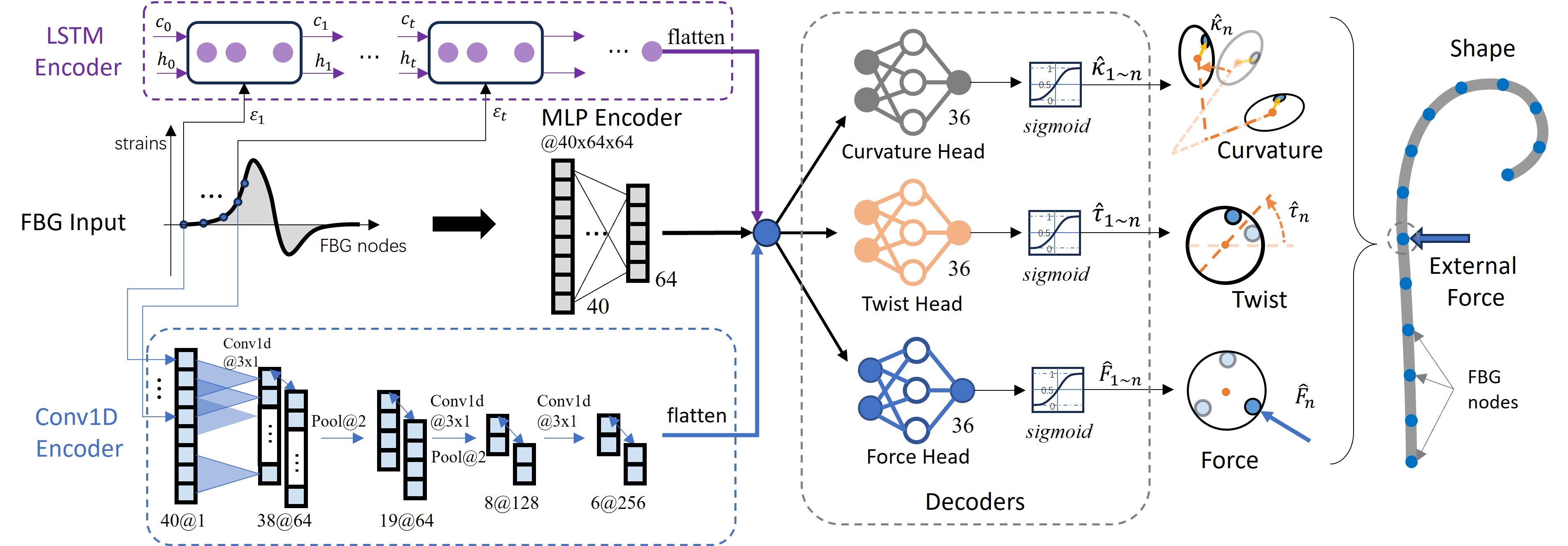}
  \caption{Overall diagram of the proposed learning models for simultaneous shape and force estimation of flexible surgical robots..}
  \label{fig:network}
  \vspace{-0.4cm}
\end{figure*}

\subsection{Design of Soft Tube with Helically Wrapped FBGs}
Flexible ureteroscopy has become a dominant way to diagnose and treat urological diseases as illustrated in Fig. \ref{fig:diagram} (a).
A flexible ureteroscope is commonly utilized during the procedures, which can bend up to 270$^\circ$ with a minimum radius of curvature ${\rm min}(r_n) = 7.5$ mm, thus it has a maximum curvature ${\rm max}(\kappa_n) \approx 133.33$ m$^{-1}$ and a deflection ratio \cite{sefati2020data} larger than 100\%.
When with tools inside the working channel of the ureteroscope (e.g., laser fiber, biopsy forceps, and stone basket), it is hard to integrate a multi-core FBG fiber inside the same channel for shape and force perception.
Consider the off-center case using FBG sensors in a straight configuration, the extension of the bending fiber outside can reach $0.5 d_e /{\rm min}(r_n) = 20 \%$ with $d_e = 3$ mm being the outer diameter of the ureteroscope, which would result in the breaking of the fiber.

Therefore, an FBG-based soft sensing tube as an off-center paradigm is proposed to be efficient and lower-cost that can simultaneously estimate the shape and force of the RAFUS.
As shown in Fig. \ref{fig:diagram} (b), we design and fabricate a highly deformable thin-walled silicone tube as the flexible substrate with a helical groove on the cylindrical surface to integrate the FBG sensors.
The tube is capable of omnidirectional bending with an inner diameter of 3 mm to assemble the ureteroscope insides, and an outer diameter $d_t = 4$ mm allowing enough wall thickness to embed the FBG fiber and smooth insertion and retraction of the tube within the flexible sheath during urology procedures.
A single-core FBG fiber is helically wrapped and adhered inside the helical groove using silicone adhesive with high tensile and shear strengths to satisfy the required deformation.
We first consider the bending radius of FBG fiber $R_f$ and determine the pitch of each helix $h_s = 30$ mm by following
\begin{equation}
\begin{aligned}
    \label{eq:pitch}
    R_f =
    (r_t^2 + (h_s/2 \pi)^2)/r_t
\end{aligned}
\end{equation}
where $r_t = d_t / 2$. 
This bending radius $R_f$ should be larger than the minimum fiber bending radius of 6 mm to guarantee the reflectivity and prevent signal distortion leading to significant noises.
Note that a smaller pitch $h_s$ can be selected to increase the spatial resolution.
Then we can determine the fiber length and total number of FBGs $N$ to be used.
Overall, we fabricate and assemble a soft sensing tube with a 120 mm length and four helices to cover the entire distal continuum mechanism of the ureteroscope.


\subsection{FBG Sensing in Helical Configuration}
Employing the proposed soft sensing tube, a model-based shape sensing method based on a single-core FBG fiber in helical configuration is introduced for comparison with the learning models in Section III.
Along the helical fiber, we can measure each FBG strain through the Bragg wavelength $\lambda$ and its shift $\Delta \lambda$ with constant temperature:
\begin{equation}
\begin{aligned}
    \label{eq:lambda}
    \varepsilon =
    \frac{\Delta \lambda}{\lambda (1 - p_{\varepsilon})}
\end{aligned}
\end{equation}
where $p_{\varepsilon}$ denotes the strain coefficient.
At each FBG section as shown in Fig. \ref{fig:diagram} (c), the FBG strain $\varepsilon_s, s \in \{1, 2, \dots, M\}$, can be decomposed into an axial strain $\varepsilon_{v,s}$ and a shear strain $\varepsilon_{h,s}$ in the orthogonal (vertical and horizontal) directions \cite{xu2016curvature}.
Then we formulate a strain-curvature-twist model mapping the axial strain $\varepsilon_{v,s}$ and shear strain $\varepsilon_{h,s}$ to the curvature $\kappa_s$ and twist $\phi_s$ with its change $\Delta \phi_s$ as 
\begin{equation}
\begin{aligned}
    \label{eq:epsilon}
    & \varepsilon_{v,s} = -\kappa_s r_s sin(\phi_s + \alpha_s) \\
    & \Delta \phi_s = \varepsilon_{h,s} \frac{\Delta h}{r_s g_{\varepsilon}}
\end{aligned}
\end{equation}
where $r_s$ and $\alpha_s$ represent the position and orientation of the FBG sensor, respectively, $\Delta h$ is the difference in arc length between two overlaid FBG sections, and $g_{\varepsilon}$ denotes a material coefficient of shear strain \cite{khan2021curvature}.

Unlike those using multi-core FBGs, only one FBG is located at each section in this work, so we group each three consecutive FBGs along the fiber with the corresponding nonlinear mappings (\ref{eq:epsilon}) as one set, to determine the curvature $\kappa_s$ and twist $\phi_s$ of these three FBG sections.
The accuracy of this model-based solution may be sacrificed due to the limited spatial resolution.
A strain bias $\varepsilon_{0,s}$ resulted from common strain and change of temperature can be taken into consideration \cite{sefati2020data}.
Consequently, the 3-D shape represented by a sequence of Cartesian positions $\bm{p}_s = \begin{bmatrix} x_s & y_s & z_s \end{bmatrix}^{\intercal} \in \mathbb{R}^{3}$, $s \in \{1, 2, \dots, M\}$, can be reconstructed using the curvatures and twists from proximal (base) to distal (tip) along the fiber based on homogeneous transformations, Frenet-Serret frame, or Bishop frame \cite{khan2021curvature, donder2021kalman}.
Note that the spatial resolution of this model-based method could be increased by using three helical fibers \cite{xu2016curvature}, but it would increase the cost and assembly complexity.
As a comparison, we will introduce the data-driven methods by using sparse strain measurement from the single-core FBG fiber in Section III.



\section{Simultaneous Estimation of Shape and Force}
By using sparse strain measurement from the proposed FBG-based soft tube, this section introduces learning-based models based om three classical encoders to achieve simultaneous shape-force sensing. As shown in Fig. \ref{fig:network}, the networks consists of a shared spatial FBG encoder and multiple heads as decoders, i.e., a) curvature/twist heads for recovering the shape, and b) force estimation head for regressing external force magnitude and contact location. 
\subsection{Spatial FBG Strain Encoder}
To ensure the robustness of the network against scale/range deviation of FBG strains $\varepsilon_{s}$, we apply normalization to process each $\varepsilon_{s}$ into the standard distribution $N(1,0)$: 
\begin{equation}
    \hat{\varepsilon} = ({\varepsilon} - E[\varepsilon])/{Std[\varepsilon]}
\end{equation}
where $E[\cdot]\in \mathbb{R}^{M}$ and $Std[\cdot]\in \mathbb{R}^{M}$ is the expected mean and standard for each local FBG node strains, and the normalized strains $\hat{\varepsilon}$ is the input of the strain encoder. 

Here we propose and evaluate three different encoders for encoding FBG strain features based on a) fully connected (FC), b) LSTM, and c) Conv1D layers.
While the FC is trained with the individual strains without considering the shape continuity, i.e., the adjacency of positions, the last two encoders are used to extract the interior spatial characteristics between neighbor FBG nodes in different manners.

\subsubsection{FC} 
FC layers are commonly employed to fit highly nonlinear relationships, and can be trained end-to-end using back-propagation. Denote the neural unit layer with the weight $W$ and bias $b$. Using normalized FBG strains $\hat{\varepsilon}$, we calculate the activation $E$ in the subsequent hidden layers:
\begin{equation}
    E^1 = f(W \cdot \hat{\varepsilon} + b), \quad
    E^l = f(W \cdot E^{l-1} + b)
\end{equation}
where $f$ is the rectified linear unit (ReLU) function, $l$ is the layer index.
The FC encoder we construct consists of a 40-neuron input layer and two 64-neuron hidden layers.

\subsubsection{LSTM} 
Different from the conventional LSTM models in encoding the time-domain sequence data \cite{li2022learning}, we convert it into extracting the spatial features between neighbor FBG nodes to generate a more accurate prediction.
Specifically, we input the sequence of FBG strains $\hat{\bm{\varepsilon}}_{n,k}=\{\varepsilon_1, \varepsilon_2, \cdots, \varepsilon_{M}\} \in \mathbb{R}^{M}$ from proximal (base) to distal (tip) into the LSTM unit in a continuous manner, as shown in Fig. \ref{fig:network}. The interior calculation is listed as follows:
\begin{equation}
    (c_t, h_t) = LSTM(\varepsilon_s,(c_{t-1}, h_{t-1}))
\end{equation}
where $\varepsilon_s$ is the FBG strain input, $c$ and $h$ are the cell input activation vector and the hidden state/output of LSTM unit, respectively. 
Here we opt for an LSTM unit with a hidden size of 64 and a stack of 3 layers. 
The final outputs $h$ are flatten to serve as inputs for the subsequent decoders.

\subsubsection{Conv1D}
Inspired by the Convolutional Neural Networks (CNN), which have been proven effective for image-based tasks using Conv2D layers to encode 2D pixel spatial features, we apply this idea to our 1D sequence FBG strain data.
In our specific scenario where deformation and FBG nodes are positioned along a 1D sequence and node positions significantly impact the overall shape (e.g., errors accumulating from proximal to distal), we employ Conv1D layers to capture the spatial relationships among neighboring FBG nodes, enhancing the accuracy of shape reconstruction.
We follow previous notions and utilize a sliding window technique to encode FBG features:
\begin{equation}
    out^{i}_{ch} = b_{ch} + \sum_{k=0}^{C-1} W_{ch} \cdot in_k
\end{equation}
where $out^{i}_{ch}$ is the $i$-th output in the channel $ch$ with $in_k$ being the input, $b_{ch}$ and $W_{ch}$ are the bias and weight vectors of Conv1D layer, respectively.
As depicted in Fig. \ref{fig:network}, we first use a Conv1D layer with in-channels of 1 and out-channels of 64 for encoding FBG features, employing the ReLU activation function.
Then we downsample these features using max-pooling with a size of 2.
We repeat the above operations using extra Conv1D layers with 128 and 256 out-channels to extract the deeper features.
The resulting features are then flattened and input to the decoders.

\begin{figure}[!ht]
  \centering
  \includegraphics[width=\linewidth]{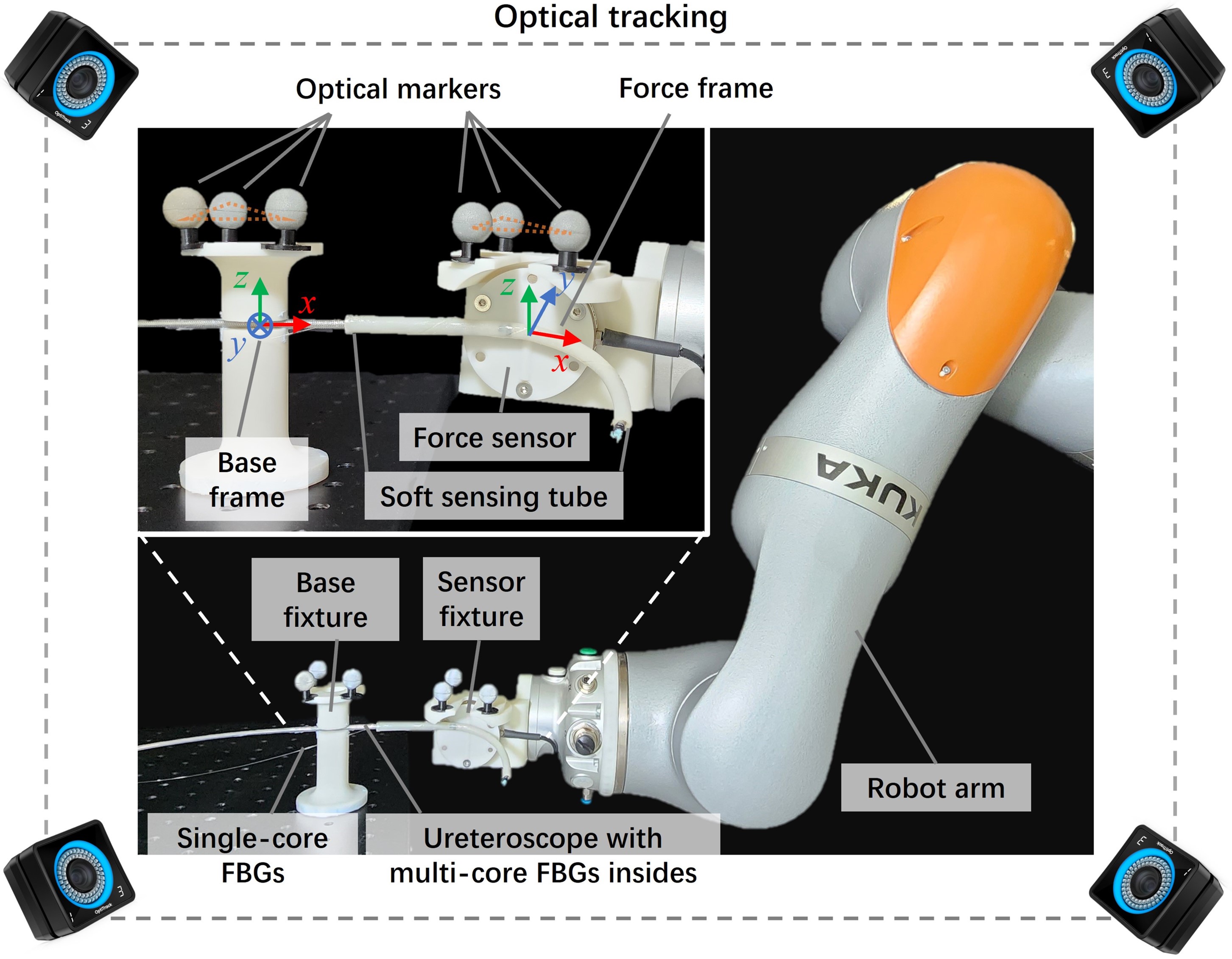}
  \caption{Data collection platform including a helical single-core FBG fiber, a multi-core FBG fiber, and a force-optical tracking system.}
  \label{fig:platform}
  \vspace{-0.4cm}
\end{figure}

\subsection{Shape Sensing Head} 
As mentioned in Section II-B, the shape can be reconstructed uniquely based on the curvature and corresponding twist of FBG nodes. 
Therefore, we establish separate curvature and twist heads as the shape decoder.
To accommodate varied scales of curvature/twist features, we initially preprocess the ground truth (GT) curvature/twist by applying a normalized mapping that scales the minimum/maximum values within the workspace to 0/1, respectively.
The sigmoid function is employed to map the predictions into [0,1].
We use the combined loss of curvature/twist for shape reconstruction based on Mean Square Error (MSE):
\begin{equation}
    L_{shape} = \|\kappa_s - \hat{\kappa}_s\|_2 + \|\phi_s - \hat{\phi}_s\|_2
\end{equation}
where $\hat{\phi}_s$ and $\hat{\kappa}_s$ represent prediction corresponding to the GT $\phi_s$ and $\kappa_s$.
During inference, we further re-scale the raw prediction into the target value from the corresponding min/max value: $y = \hat{y} \cdot (y_{max} - y_{min}) + y_{min}, y\in\{\kappa_s,\phi_s\}$.

\subsection{Force Estimation Head} 
This task involves recovering external force magnitude and contact location from raw FBG sensor data.
To overcome the inability learning the process caused by the discontinuity of force data, we convert the impulse form of the force along the shape into a continuous distribution as follows:
\begin{equation}
    f_i = F_c * exp^{(x_i - x_c) ^ 2} / (2 \cdot \sigma_f^2), \quad i\in\{1, 2, ...\}
\end{equation}
where $f_i$ is the force at location $x_i$ caused by external force with magnitude $F_c$ at location $x_c$. 
$\sigma_f$ is the operation range variance.
For force-related training, the force loss is the MSE on the overall distribution: $L_{force} = \sum_i \| f_i - \hat{f}_i\|_2$.
During inference, the force head simultaneously predicts the overall force distribution and the corresponding magnitude.
The contact location can be obtained accordingly using $\hat{x}_c=\argmax_{i\in\{1,2,\cdots\}} F_i$.
The force magnitude is also re-scaled into the normal range from the min/max force.

\section{Experiments}
To evaluate the performances of the proposed methods in both free and constrained environments, we introduce the experimental setup and discuss the results in this section.

\begin{table}[!ht]
\caption{\centering Comparison results of static experiments}
\scriptsize
\label{table:static}
\centering
\begin{tabular}{|c|c|c|c|c|}
\hline
       & \begin{tabular}[c]{@{}c@{}}Tip position\\ error (mm)\end{tabular} & \begin{tabular}[c]{@{}c@{}}Shape\\ error (mm)\end{tabular} & \begin{tabular}[c]{@{}c@{}}Force magnitude\\ error (mN)\end{tabular} & \begin{tabular}[c]{@{}c@{}}Force location\\ error (mm)\end{tabular} \\ \hline
FC     & 1.82 $\pm$ 1.94                                             & 0.67 $\pm$ 0.55                                      & 40.69 $\pm$ 41.51                                              & 2.58 $\pm$ 4.28                                               \\ \hline
LSTM   & 1.56 $\pm$ 1.66                                             & 0.50 $\pm$ 0.41                                      & 37.59 $\pm$ 40.84                                              & 2.67 $\pm$ 4.19                                               \\ \hline
Conv1D & 0.93 $\pm$ 1.03                                             & 0.35 $\pm$ 0.31                                      & 35.25 $\pm$ 38.34                                              & 2.35 $\pm$ 3.38                                               \\ \hline
\end{tabular}
\vspace{-0.4cm}
\end{table}

\subsection{Setup}
We integrated a platform to conduct the experiments, including an RAFUS system, FBG sensing units, a force-optical tracking system, and a robot arm.
The RAFUS system was designed and assembled, including a single-use disposable flexible ureteroscope (UR-F1, BASDATA) equipped with the FBG-based soft sensing tube, and actuation units.
In addition to the parameters of the soft sensing tube introduced in Section II, its helical groove is 0.3 mm deep and 0.4 mm wide to secure a single-core FBG fiber.
The single-core fiber (FBGS) with 14 FBGs, 195 $\mu$m outer diameter, and 132 mm length, was wrapped and adhered into the groove using bio-compatible silicone adhesive (S-8005N, ALDERS), which can provide similar elastic modulus with the soft tube.

\begin{figure}[!ht]
  \centering
  \includegraphics[width=\linewidth]{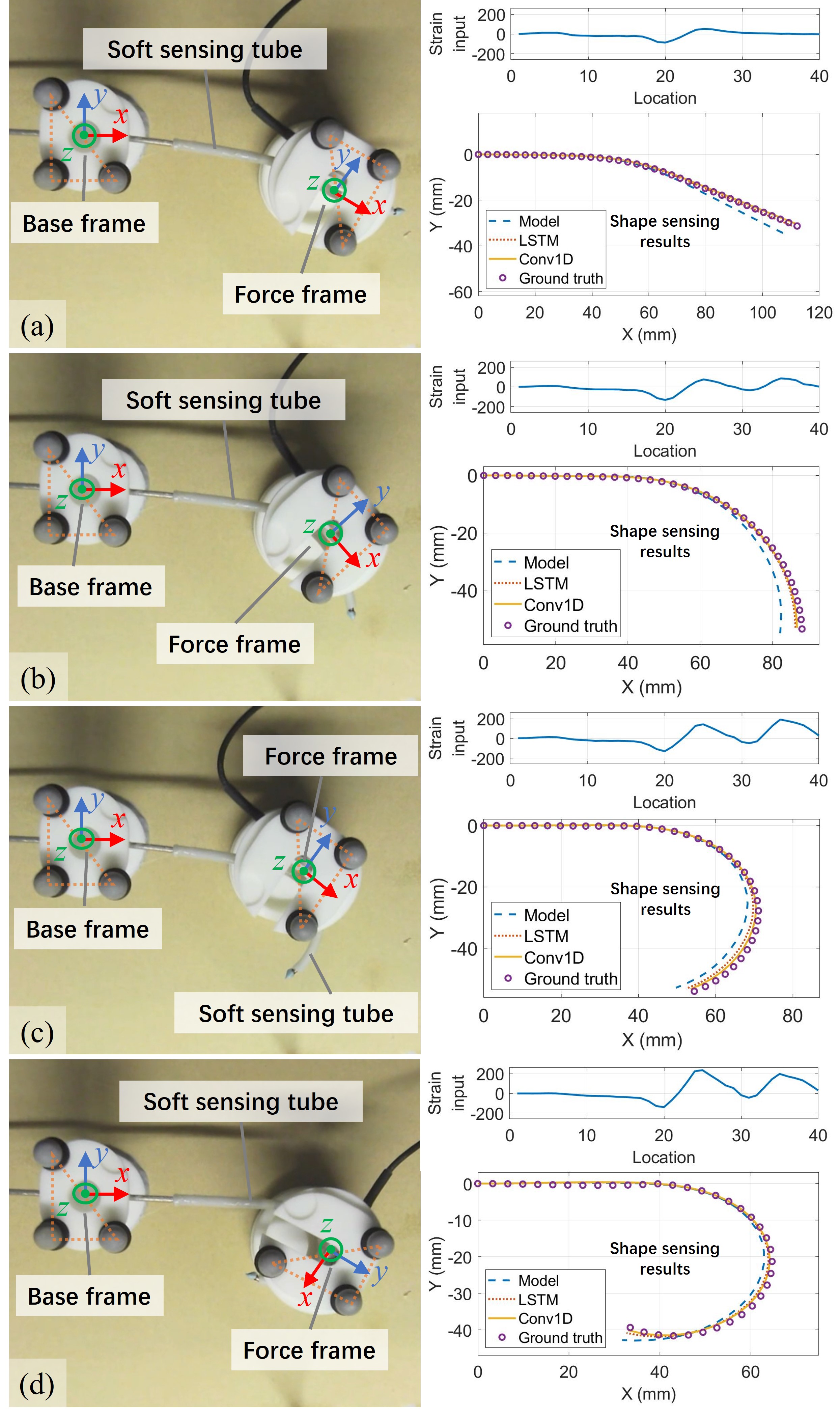}
  \caption{Task setups with initial bending (left) of (a) 30$^\circ$, (b) 85$^\circ$, (c) 120$^\circ$, and (d) 185$^\circ$, and their strain input from helical single-core FBGs (top-right) with shape sensing results (bottom-right).}
  \label{fig:setup}
  \vspace{-0.4cm}
\end{figure}

\begin{figure*}[!ht]
  \centering
  \includegraphics[width=\linewidth]{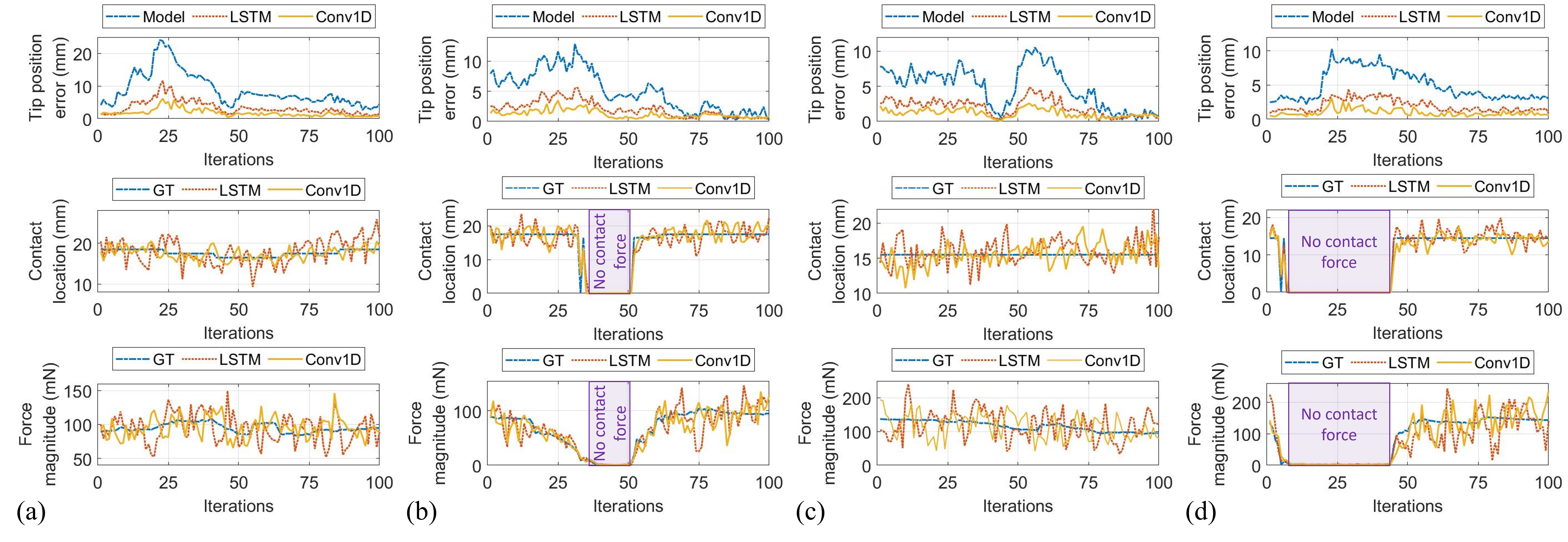}
  \setlength{\abovecaptionskip}{-0.4cm}
  \caption{The results versus time iteration including tip position error (top), contact location (middle), and force magnitude (bottom) of four experiments (a) - (d) corresponding to the tasks in Fig. \ref{fig:setup} (a) - (d), respectively. Note that the middle and bottom rows show the ground truth (GT) in blue dash-dotted curves instead of the results using the model-based method (Model) in the top row.} 
  \label{fig:exp1}
  \vspace{-0.35cm}
\end{figure*}


For the proposed learning models, the curvature/twist/force heads have the same architecture of the one fully connected layer, which is empirically sufficient to decode the features from the previous encoders.
All the network models were trained with a learning rate of 2e-3, and a batch size of 256.
We used Adam optimizer with a 1e-5 weight decay. 
6224 samples were collected, of which 80\% were selected for training and the rest were used for testing.
$\sigma_{f}=3$ was used in the force estimation head.


\subsection{Data Collection}
To collect the data for training proposed network models, we integrated a platform as illustrated in Fig. \ref{fig:platform}.
It consists of a soft sensing tube with a helically wound single-core FBG fiber, a multi-core FBG fiber, and a force-optical tracking system.
The spatial resolution of FBG fibers was 3.3 mm, hence totally $M = 40$ strains were measured along the single-core fiber as the input.
We embedded a four-core fiber with draw tower gratings (DTGs, FBGS) inside the ureteroscope working channel, together with a robust and accurate sensing algorithm \cite{lu2023robust} to acquire the GT curvatures for shape reconstruction of the ureteroscope. 
The strain signals from both single-core and multi-core fibers were simultaneously decoded via an OFDR interrogator (RTS 125+, Sensuron).
Note that it is feasible to implement the proposed methods with an interrogator based on wavelength-division multiplexing (WDM) as a lower-cost alternative.

For force sensing validations, we fabricated a force-optical tracking unit as shown in Fig. \ref{fig:platform}.
The GT force magnitude was measured via a force sensor (Nano43, ATI), which was attached to a 3-D printed sensor fixture with optical markers.
From optical cameras (Prime 13, OptiTrack), we could locate the GT contact position by tracking the pose of the force frame on the sensor fixture related to the base frame on another base fixture.
Besides, a robot arm (LBR iiwa, KUKA) was deployed to position the sensor fixture, thus enabling the force sensor contact with the ureteroscope for efficient data collection.

\subsection{Results and Discussion}
Experiments comparing the model-based sensing approach with the learning-based methods introduced in this paper, were performed in static and dynamic conditions, respectively.
Tip position, shape, force magnitude, and contact location were recorded as four metrics, and their errors were calculated, which are accordingly presented in TABLE \ref{table:static}, \ref{table:dynamic}, and Fig. \ref{fig:exp1}.
We define the Euclidean distance from the estimated tip Cartesian position to its GT as the tip position error.
The average position errors of all sensing points 
along the robot is recorded as the shape error.

\subsubsection{Static experiments}
Static experiments were first performed to validate the proposed learning models after training. 
We tested 1224 cases with different bending angles from -270$^\circ$ to 270$^\circ$, contact locations along the soft tube within a 90 mm length, and force magnitudes in a range of 0 $\sim$ 500 mN.
The results in terms of four metrics using three network models are listed in TABLE \ref{table:static}.
We can observe that the Conv1D network outputs the smallest errors of all metrics, which is contributed by its consideration of shape continuity in the spatial strain encoder.
Specifically, as compared to the average tip position errors of 1.82 mm (1.52\% of the total length) and 1.56 mm (1.30\%) using the FC and LSTM networks, respectively, the error of the Conv1D model is reduced by nearly two times as 0.93 mm (0.78\%). 

\subsubsection{Dynamic tracking experiments}
Based on the results of static experiments, dynamic tests were performed to compare the model-based approach with the LSTM and Conv1D models in 20 kinds of setups, and four of them are illustrated in Fig. \ref{fig:setup}.
The differences among these setups are similar to those in the static experiments concerning bending angles as the initial configurations, contact locations, and force magnitude.
Moreover, the flexible robot bent from the initial configurations, e.g., 30$^\circ$, 85$^\circ$, 120$^\circ$, and 185$^\circ$ in contact with the sensor fixture as indicated in the right figures of Fig. \ref{fig:setup} (a) - (d), respectively, to 270$^\circ$, and then bent back to the initial configurations. 
The top-right plots in Fig. \ref{fig:setup} demonstrate the corresponding strain inputs from the FBG-based soft tube, and the bottom-right plots show the shape sensing results of the model-based, LSTM, and Conv1D methods, respectively, in blue dash-dotted, red dotted, and yellow solid curves, in comparison with the GT in purple circles.

The results varying versus time iteration are plotted in Fig. \ref{fig:exp1} corresponding to the four setups in Fig. \ref{fig:setup}, where tip position error, contact location, and force magnitude are shown from top to bottom.
Since the model-based method is introduced for only shape sensing without force estimation, GT contact location and force magnitude are presented in blue dash-dotted curves of Fig. \ref{fig:exp1} instead.
The tip position errors using the model-based approach are larger than those of the two networks.
Besides, although the robot was in contact with the sensor fixture initially, it would become without contact when approaching 270$^\circ$ bending angle as depicted by purple solid boxes in Fig. \ref{fig:exp1} (b) and (d), where the contact locations are marked as 0 and so are the force magnitudes.
The results are also summarized in TABLE \ref{table:dynamic}, from which two networks outperform the model-based one regarding all metrics with the smallest average errors of tip position as 1.05 mm (0.88\%), force magnitude as 27.81 mN (12.39\%), 
and force location as 3.79 mm (3.16\%), which are comparable with the state-of-the-art studies \cite{qiao2021force,ha2022contact}. 
Furthermore, most results using Conv1D slightly outperform LSTM because long-term memory may not contribute to the spatial continuity.
These results support that our learning paradigm can alleviate precise FBG placement as required by the conventional method, and it has better accuracy and robustness in dynamic conditions with unknown external interactions.
A limitation of our method is that it was only validated with 2-D deformation with in-plane bending.
Therefore, a data collection platform with measurements in 3-D space will be developed to further evaluate the higher dimensional sensing capability.


\begin{table}[]
\caption{\centering Comparison results of dynamic experiments}
\scriptsize
\label{table:dynamic}
\centering
\begin{tabular}{|c|c|c|c|c|}
\hline
       & \begin{tabular}[c]{@{}c@{}}Tip position\\ error (mm)\end{tabular} & \begin{tabular}[c]{@{}c@{}}Shape\\error  (mm)\end{tabular} & \begin{tabular}[c]{@{}c@{}}Force magnitude\\error  (mN)\end{tabular} & \begin{tabular}[c]{@{}c@{}}Force location\\error  (mm)\end{tabular} \\ \hline
Model  & 5.00 $\pm$ 3.37                                             & 1.55 $\pm$ 0.79                                      & NA                                              
            & NA                                               \\ \hline
LSTM   & 1.71 $\pm$ 0.85                                             & 0.66 $\pm$ 0.24                                      & 35.26 $\pm$ 24.18                                              & 5.54 $\pm$ 4.35                                               \\ \hline
Conv1D & 1.05 $\pm$ 0.59                                             & 0.48 $\pm$ 0.14                                      & 27.81 $\pm$ 19.54                                              & 3.79 $\pm$ 3.25                                               \\ \hline
\end{tabular}
\vspace{-0.4cm}
\end{table}

\section{Conclusion}

This paper proposes a novel data-driven approach for simultaneous shape-force estimation along highly deformable flexible robots.
We design a disposable and low-cost soft sensing tube with a helical single-core FBG fiber, and study three learning models based on sparse strain measurement.
The proposed methods have been thoroughly validated in free space without interactions and constrained environments with contact forces, showing their effectiveness and superiority.

In the future, we will investigate this paradigm in fusion with self-contained actuation input and eye-in-hand camera vision for robotic flexible ureteroscopy and laparoscopy \cite{wang2023barrier,li2024gmm}.
The temporal characteristics of the flexible robots (e.g., hysteresis) and other factors (e.g., temperature and humidity) will be considered to improve the performance.


\newpage

\addtolength{\textheight}{-12cm}   

\bibliographystyle{IEEEtran}
\bibliography{./references}

\end{document}